\documentclass[journal]{IEEEtran}
\ifCLASSINFOpdf
\else
\fi
\usepackage{mathtools}
\usepackage{color}
\usepackage{algorithm}
\usepackage[noend]{algpseudocode}

\usepackage{times}
\usepackage{epsfig}
\usepackage{graphicx}
\usepackage{amssymb}
\usepackage{romannum}
\usepackage{color}

\usepackage{amsmath}
\usepackage{breqn}

\begin{document}
%
\title{Ensemble BERT for Medication Event Classification on Electronic Health Records (EHRs)}
%
%
%

 \author{Shouvon Sarker, Xishuang Dong,  and~Lijun Qian,~\IEEEmembership{Senior Member,~IEEE}
\thanks{X. Dong, S.  Sarker, and L. Qian are with the Center of Excellence in Research and Education for Big Military Data Intelligence (CREDIT Center), Department of Electrical and Computer Engineering, Prairie View A\&M University, Texas A\&M University System, Prairie View, TX 77446, USA. Email: xidong@pvamu.edu, ssarker3@pvamu.edu, liqian@pvamu.edu}
}

\maketitle

\begin{abstract}
Identification of key variables such as medications, diseases, relations from health records and clinical notes has a wide range of applications in the clinical domain. 
n2c2 2022 provided shared tasks on challenges in natural language processing for clinical data analytics on electronic health records (EHR), where it built a comprehensive annotated clinical data \textit{Contextualized Medication Event Dataset (CMED)}. 
This study focuses on subtask 2 in Track 1 of this challenge that is to detect and classify medication events from clinical notes through building a novel BERT-based ensemble model. It started with pretraining BERT models on different types of big data such as Wikipedia and MIMIC. Afterwards, these pretrained BERT models were fine-tuned on \textit{CMED} training data. These fine-tuned BERT models were employed to accomplish medication event classification on \textit{CMED} testing data with multiple predictions. These multiple predictions generated by these fine-tuned BERT models were integrated to build final prediction with voting strategies.  
Experimental results demonstrated that BERT-based ensemble models can effectively improve strict Micro-F score by about 5\% and strict Macro-F score by about 6\%, respectively. 
 \end{abstract}


%
\IEEEpeerreviewmaketitle

\section{Introduction}
\label{sec1}
Electronic Health Records (EHRs) mostly contain unstructured texts regarding patient’s history and health conditions, discharge paper and clinical notes. The unstructured texts make it very difficult to process for most state-of-the-art language model representations\cite{GoodBengCour2016}. Also, the scarcity of biomedical dataset related to the privacy and the security of the patients make it very difficult to train a deep learning model. To promote the development of data analytics on EHRs, n2c2 competitions organized by Harvard Medical School recruit teams from the world to complete various tasks such as named entity recognition (NER), relation extraction (RE) and question answering (QA). It released 500 EHRs with annotations as Contextualized Medication Event Dataset (CMED) for the tasks related to medication identification and event classification\cite{cmed}, where the event recognition aims to detect and classify medication and associated medication changes mentioned in the clinical notes into Disposition (if there is a change of medication mentioned), NoDisposition (if there is no change of medication mentioned) and Undetermined (if more information is required). 

BERT (Bidirectional Encoder Representation from Transformers)\cite{bert} is proposed by Google to complete various text-ming tasks. Unfortunately, BERT are not able to perform information extraction on EHRs due to the variants of medical terms in the EHRs, which needs language models that are specifically pre-trained on the clinical text domain. In 2019, BioBERT was introduced as a language representation model specifically pre-trained on clinical domain\cite{biobert}. It was shown that pre-trained BERT on biomedical domain outperforms BERT in various biomedical text mining tasks such as named entity recognition, relation extraction and question answering. Later some other versions of BioBERT such as clinical BERT that is also pre-trained on clinical notes were introduced. The overall process of pre-training and fine-tuning BioBERT consists of two steps. First the weight from BERT which was pre-trained on general domain corpora (English Wikipedia and Book corpus) was initialized. Then it is pre-trained on PubMed abstracts and PMC full text articles. 

This paper proposed a novel ensemble model based on BERT to enhance performance of medication event classification. It included four steps: 1) Pretraining BERT models on different types of big data such as Wikipedia and MIMIC; 2) Fine-tuning pretrained BERT models on \textit{CMED} training data; 3) Applying these fine-tuned BERT models to perform medication event classification on \textit{CMED} testing data and generating  multiple predictions;  4) Integrating these multiple predictions with voting strategies to generate final predictions. Moreover, different post-processing strategies were explored to manipulate the final predictions to extract medications from the \textit{CMED} testing data. Experimental results demonstrated that the proposed ensemble model can enhance the performance of the medication event classification effectively, as well as obtain promising performance for medication extraction.

The rest of this paper is organized as: in section 2, it presented the proposed method in detailed; Section 3 showed and discussed the experimental results in detail. Section 4  and 5 presented detailed related work and conclusions, respectively.

\section{Methodology}
\label{sec4}


Since the release of BERT in 2018, it has been the benchmark for most of the text mining tasks. While BERT and other sequential models such as Recurrent Neural Networks (RNN) and Long Short-term Memory (LSTM) have been able to achieve encouraging performance in the general language domain, there has not been yet any benchmark model for the biomedical domain\cite{overview}\cite{survey}\cite{out}.

\subsection{BERT Model}
In this work, BERT was used as the baseline for our work\cite{transformer}. BERT can take up to a sequence length of 512 tokens. Using 512 tokens per sequence is very computation costly. Before feeding the tokens of the sequence into the model, BERT normally replaces 15\% of the tokens in each sequence by a [Mask] token. After the tokens are fed into the model, BERT tries to predict of a masked token based on content provided by the non-masked tokens of the sequence. BERT adds a classification layer on top of the encoder. It multiplies the output vectors by the embedding 14 matrix\cite{vec}. Therefore, they are transformed into a vocabulary dimension. Then it calculates the probability of each word in the vocabulary\cite{bert}.

In the training, BERT receives a pair of sequences. BERT tries to predict if the second sequence of the pair is sequence from the same document. In time of training, 50\% of the sentences are from the same documents and the 50\% documents are chosen over the random documents as the second sequence. It is assumed that in time of prediction, the second sequence will be separated from the first sequence \cite{word}. In order to differentiate the two sentences, BERT normally inserts a [CLS] token at the beginning of the sequence and a [SEP] token at the end of the sequence. These two [CLS] and [SEP] tokens help BERT to distinguish between the sentences\cite{bert}. However, BERT does not perform well when it comes to clinical and biomedical texts because the general vocabulary representation and distribution in the general domain can be different than that in the clinical field. BERT needs to be pre-trained on the clinical data\cite{bert, overview}. 

%
%


\subsection{Ensemble BERT Models}
\label{sec:dla}
\begin{figure} [ht]
 	\centering
	\includegraphics[width=.9\linewidth]{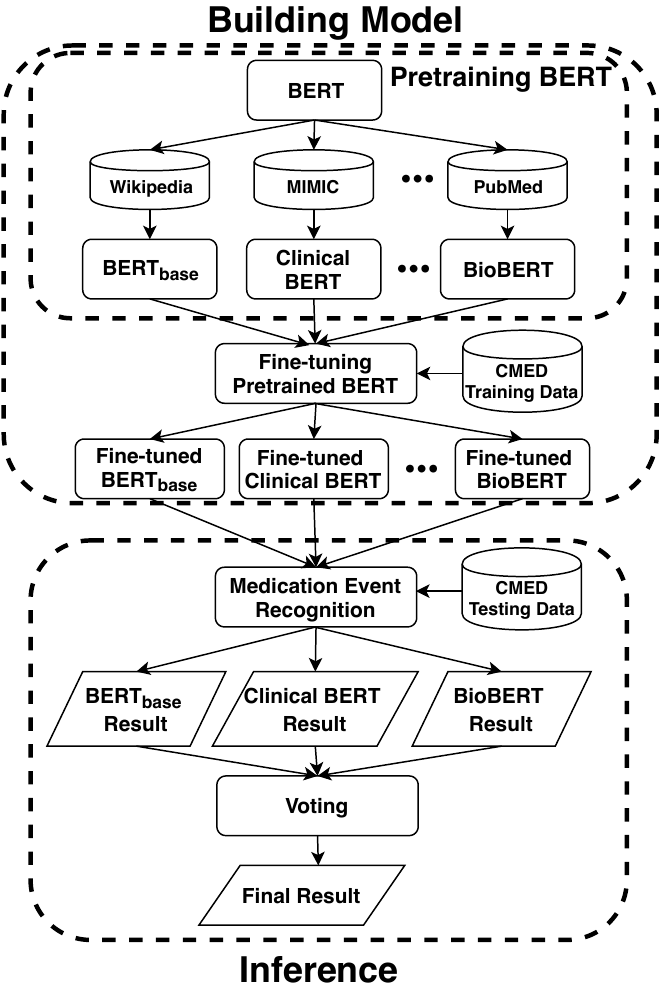}
	\caption{Flow of building ensemble BERT models }
	\label{Fig1_flow}
\end{figure}

 A voting ensemble works by combining the predictions from different models. The advantage of using ensemble models is that poor performance or wrong predictions from one model can be offset by a high performance from another model\cite{ensemble}. Voting ensemble can be used for either regression or for classification. In case of classification, the prediction from each model is summed up and then hard voting is applied to classify. Hard voting involves summing the predictions for each class label and predicting the class label with the most votes. Soft voting involves summing the predicted probabilities (or probability-like scores) for each class label and predicting the class label with the largest probability\cite{ensemble}. In the proposed method, we employed majority soft voting for the ensemble models that includes ensemble of the trained BERT models pretrained on clinical and biomedical data\cite{ensemble2}. The flow of building the proposed ensemble model and applying the model for inference is shown in Figure~\ref{Fig1_flow}.
 
In the process of building the model, it starts with pretraining BERT models on different types of data such as Wikipedia from general domain and Pubmed and MIMIC from the biomedical domains, which will produce multiple pretrained models such as $BERT_{base}$, Clinical BERT, and BioBERT.  Afterwards, these pretrained models will be fine-tuned on CMED training data for medication event classification. These fine-tuned models will be employed to perform medication event classification on CMED testing data and generate multiple recognition results. Finally, these results will be combined through applying voting to produce final results for the medication event classification. Specifically, we explored a weighted ensemble model based on expected calibration error (ECE)\cite{ece}, where the ECE values were obtained on training data. The basic rule to build this weighted ensemble model is that the models with lower ECE values are given higher weights.

\section{Experiment}
\label{sec5}

\subsection{Datasets}

Understanding the context of a real time generated electronic health record is fundamental to have a complete picture of a patient’s clinical history. In order to provide better care to a patient, health care providers must have a complete look of patient’s medication history\cite{cmed}. Most of the times, providers depend on structured medication mentions to identify changes and prescribe future medications. However, the EHRs generated at the time can be unstructured but might contain important information about patient’s medication and clinical history. Furthermore, it is a challenging task to mine the important information of medication mentions and medication event changes from the unstructured EHRs. 

Most of the previous research works have not fully investigated how to identify medication and classify medication changes in an electronic health record.  To result this issue, N2C2 2022 competition provided an annotated dataset, Contextualized Medication Event Dataset (CMED), that captures the context of medication changes in health records. It organizes medication events into orthogonal dimensions and captures relevant context\cite{cmed}. This dataset was annotated over 500 electronic health records which were randomly selected  by a team of three annotators. It consists of $9,013$ medication mentions annotated considering their contexts in the EHRs. For each medication mention, the annotator first determines if a change is being discussed (Disposition) or not (NoDisposition); if more information is needed to determine this, the mention is labeled as Undetermined.  For the competition, 80\% of the data with their annotations was released as training and evaluation data which consisted of 80\% of the data and $7,230$ medication mentions over 400 notes\cite{cmed}. 

 \begin{figure} [ht]
 	\centering
	\includegraphics[width=.9\linewidth]{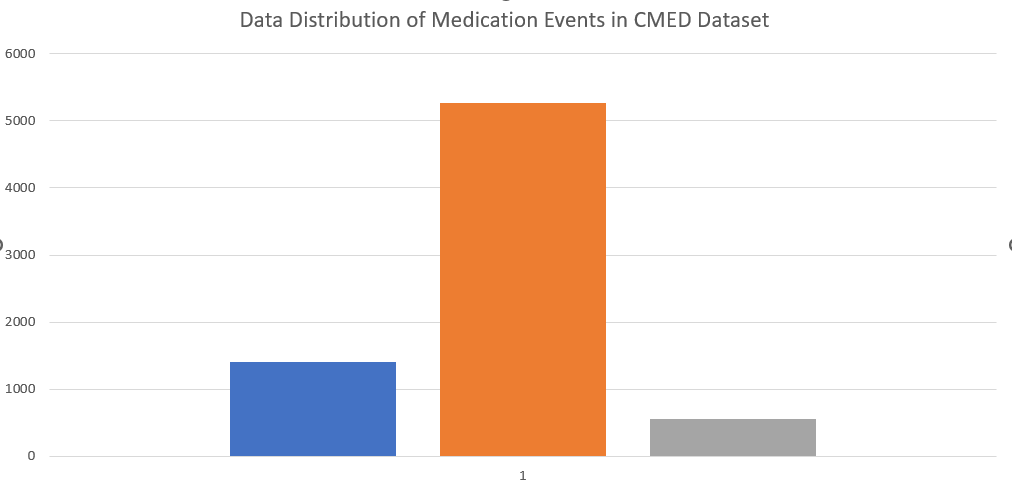}
	\caption{Data distribution among three classes}
	\label{Fig_4_nodes_data}
\end{figure}

Figure~\ref{Fig_4_nodes_data} presented data distribution of three classes, namely, Disposition, NoDisposition and Undetermined. Only 7\% of the total data belongs to the Undetermined class, thus the published CMED dataset is imbalanced. 

Tokens from the EHRs were tagged using the BIO format\cite{iob}. If there is a medication and any change of it has been discussed in the respective EHR note, it was labeled as B-Disposition or I- Disposition depending on the beginning or the inside of the chunk\cite{iob}. 
An example of tagging is shown in the below table:

\begin{table}[ht]
\caption{An example of preprocessed annotated EHRs. The original sentences were preprocessed by removing stop words.}

\centering
\begin{tabular}{|c|c|c|}
\hline
\textbf{Original Sentence} & \textbf{Preprocessed Sentence} & \textbf{Tags} \\
\hline
He is not &    He&	O\\ 
currently &currently&	O\\ 
using &using	&O\\
metronidazole& metronidazole&	B-Disposition\\
pill& pill&	O\\
any longer& longer&	O\\
\hline
\end{tabular}
\label{tab:taging}
\end{table}

If no change of the mentioned medication is discussed in EHR
note, then it was labeled as B-NoDisposition or I-NoDispositon. If more information
was required to determine a change of the medication in a note, then it was labeled as
B-Undetermined or I-Undetermined.

\subsection{Evaluation Metrics}
\label{sec:metrics}

Considering the medication event classification as a multiclass classification problem, we applied different evaluation metrics to evaluate the performance of our proposed model, which includes macro-average Precision (MacroP), macro-average Recall (MacroR), macro-average Fscore (MacroF)~\cite{van2013macro}, and micro-average Precision (MicroP), micro-average Recall (MicroR), micro-average Fscore (MicroF)~\cite{van2013macro}. Macro-average~\cite{yang2001study} is to calculate the metrics such as Precision, Recall and Fscores independently for each type of medication events and then utilize the average of these metrics.  It is to evaluate the whole performance of classifying the medication events.  
\begin{equation}
	MacroF = \frac{1}{C} \sum_{c=1}^{C} Fscore_c.
\end{equation}
\begin{equation}
	MacroP= \frac{1}{C} \sum_{c=1}^{C} Precision_c.
\end{equation}
\begin{equation}
	MacroR= \frac{1}{C} \sum_{c=1}^{C} Recall_c.
\end{equation}
where $C$ denotes the total number of medication events and $Fscore_c$, $Precision_c$, $Recall_c$ are $Fscore$, $Precision$, $Recall$ values in the $c^{th}$ medication events which are defined by

\begin{equation}
	Fscore = \frac{2 \times Precision \times Recall}{Precision + Recall}.
\end{equation}

where $Precision$ indicates precision measurement that defines the capability of a model to represent only correct medication events and $Recall$ computes the aptness to refer all corresponding correct medication events:

\begin{equation}
	Precision = \frac{TP}{TP+FP}.
\end{equation}

\begin{equation}
	Recall = \frac{TP}{TP+FN}.
\end{equation}
whereas ${TP}$ (True Positive) counts total number of medication events matched with the medication events in the types. ${FP}$ (False Positive) measures the number of recognized type does not match  the annotated medication events. ${FN}$ (False Negative) counts the number of medication events that does  not match  the predicted medication events. 
The main goal for learning from imbalanced datasets is to improve the recall without hurting the precision. However, recall and precision goals are often conflicting, since when increasing the true positive (TP) for the minority class (True), the number of false positives (FP) can also be increased; this will reduce the precision~\cite{chawla2009data}. 

In addition, strict and lenient matching was considered for calculating these evaluation metrics\cite{multi}, where the strict matching refers to the two offsets associated with the medication match exactly while the lenient matching means that there is an overlap between the two offset pairs.

\subsection{Pretrained BERT}
As shown in Figure~\ref{Fig1_flow}, it starts with pretraining of BERT models.  Table~\ref{tab:pretraining} showed data for BERT pretraining in detail.
\begin{table}[ht]
\caption{Data for pretraining BERT models. The main difference between $Roberta_{base}$  and $BERT_{base}$ is that it applied different hyper-parameters to train BERT models.}
\centering
\begin{tabular}{|l|l|}

\hline
\textbf{Model} & \textbf{Pre-training Data}\\
\hline

$BERT_{base}$ & Wikipedia and Bookscorpus	 \\ 
$BioBERT_{pmc}$ & PMC Journals\\
$BioBERT_{pubmed}$	& PubMed Abstracts \\
$BioBERT_{pubmed\_pmc}$	&PubMed Abstracts, PMC Journals \\
$Clinical BERT$	&MIMIC Notes \\
$Discharge BERT$ & MIMIC-III Discharge Notes \\
$BioClinical BERT$	 &PubMed Abstracts, PMC Journals, MIMIC \\
$BioDischarge BERT$	& PubMed Abstracts, PMC Journals, MIMIC-III  \\
$BioReddit BERT$	& PubMed Abstracts, PMC Journals, Reddit \\
$Roberta_{base}$ & Wikipedia and Bookscorpus \\
$Roberta_{large}$ & Wikipedia and Bookscorpus \\
\hline

\end{tabular}
\label{tab:pretraining}
 \end{table}

\subsection{Result and Discussion}

This paper proposed a novel ensemble model based on pretrained BERT models for medication event classification. The following sections presented detailed results on this task.

\subsubsection{Medication Event Classification}

Table~\ref{tab-med-micro} and~\ref{tab-med-macro} showed the performance comparison on medication event classification via Micro-average and Macro-average evaluation metrics between pretrained BERT models and proposed ensemble models. Generally speaking, pretrained BERT models are able to obtain promising performance based on strict and lenient evaluation results. For example, $Roberta_{large}$ could obtain 0.8003 and 0.8237 for strict and lenient Fscore, respectively, which is consistent with the previous work.

We also obtained consistent observations on performance improvement with proposed models. For example, in Table~\ref{tab-med-micro}, regarding Micro-average performance, the proposed ensemble model outperformed the baselines by 9\% and 4\% for strict MicroP and MicroF values. For lenient Micro-average evaluation, the proposed ensemble model can improve performance as well regarding MicroP and MicroF values.

For Macro-average evaluation results in Table~\ref{tab-med-macro}, the proposed ensemble model can enhance the performance on all evaluation metrics. For instance, regarding strict MacroP, MacroR, and MacroF, the performance was improved by 7\%, 5\%, and 6\%, respectively. For lenient MacroP, MacroR, and MacroF, the performance was improved by 6\%, 4\%, and 5\%, respectively. 

In addition to medication event classification, we post-processed the predictions of medication event classification to complete medication identification. For instance, if the predictions included ``B-disposition" and ``I-disposition" for two tokens, we combined these two tokens as one prediction for ``drug" medication. Next section will present the performance of medication identification in detailed.

\begin{table*}[!ht]
\caption{Performance comparison on medication event classification via Micro-average evaluation metrics between pretrained BERT models and proposed ensemble models}
\centering
\begin{tabular}{|l|ccc|ccc|}
\hline
\textbf{Model}&\multicolumn{3}{c|}{\textbf{Strict Evaluation Performance}}&\multicolumn{3}{c|}{\textbf{Lenient Evaluation Performance}} \\
\hline
\textbf{Pretrained BERT}  & \textbf{MicroP} & \textbf{MicroR} & \textbf{MicroF} & \textbf{MicroP} & \textbf{MicroR} & \textbf{MicroF} \\
\hline
$BERT_{base}$	&0.7297&	0.6776&	0.7027 &	 0.7572&	0.7031&	0.7291\\
$BioBERT_{pmc}$	&0.7382	&0.6901	&0.7133 & 0.7670	&0.7170 &0.7420\\
$BioBERT_{pubmed}$&	0.7399	&0.7072	&0.7207 & 0.7637&	0.7252	&0.7440\\
$BioBERT_{pubmed\_pmc}$	&0.7381	&0.6944&	0.7156 & 0.7619&	0.7271	&0.7483\\
$Clinical BERT$	&0.7453&	0.7199&	0.7322 &0.7682	&0.7416	&0.7547\\
$Discharge BERT$	&0.7369	&0.6986	&0.7169 &0.7642	&0.7252	&0.7442\\
$BioClinical BERT$	&0.7396	&0.7188	&0.7317 &0.7620	&0.7348	&0.7482\\
$BioDischarge BERT$	&0.7296	&0.7048	&0.7170 &0.7607&	0.7358&	0.7476\\
$BioReddit BERT$	&0.7556&	0.8238	&0.7817 &0.7758	&0.8567	&0.8129\\
$Roberta_{base}$	&0.7461	&0.8210	&0.7818 &0.7719	&0.8493	&0.8087\\
$Roberta_{large}$	&0.7678	&0.8357&	0.8003 &0.7902	&0.8601	&0.8237\\
\hline
\hline
\textbf{Weighted Voting-based Ensemble}	&0.7561	&0.8255	&0.7893&0.7805	&0.8521&	0.8147\\
\textbf{Majority Voting-based Ensemble}	&0.8720	&0.8104	&0.8401 &0.8924	&0.8294	&0.8597\\
\hline
\end{tabular}
\label{tab-med-micro}
\end{table*}

\begin{table*}[!ht]
\caption{Performance comparison on medication event classification via Macro-average evaluation metrics  between pretrained BERT models and proposed ensemble models}
\centering
\begin{tabular}{|l|ccc|ccc|}
\hline
\textbf{Model}&\multicolumn{3}{c|}{\textbf{Strict Evaluation Performance}}&\multicolumn{3}{c|}{\textbf{Lenient Evaluation Performance}} \\
\hline
\textbf{Pretrained BERT}  & \textbf{MacroP} & \textbf{MacroR} & \textbf{MacroF} & \textbf{MacroP} & \textbf{MacroR} & \textbf{MacroF} \\
\hline
$BERT_{base}$	&0.6344	&0.5960&	0.6145 &	 0.6491&	0.6097&	0.6287\\
$BioBERT_{pmc}$	&0.6835& 0.6011&	0.6349 & 0.6989	&0.6146&	0.6500\\
$BioBERT_{pubmed}$&	0.6559&	0.6348&	0.6436 & 0.6696	&0.6471&0.6570\\
$BioBERT_{pubmed\_pmc}$	&0.6549	&0.6073	&0.6298 & 0.6723	&0.6237&	0.6467\\
$Clinical BERT$	&0.6830	&0.6887&	0.6843 &0.6943	&0.6993&	0.6956\\
$Discharge BERT$	&0.6633&	0.6479&	0.6547 &0.6775	&0.6613	&0.6685\\
$BioClinical BERT$	&0.6596	&0.6812	&0.6687 &0.6750	&0.6960	&0.6838\\
$BioDischarge BERT$	&0.6540	&0.6703&	0.6610 &0.6725	&0.6884&	0.6793\\
$BioReddit BERT$	&0.6774	&0.6911	&0.6816 &0.6925	&0.7096&	0.6976\\
$Roberta_{base}$	&0.6737	&0.7235	&0.6975 &0.6880	&0.7393	&0.7126\\
$Roberta_{large}$	&0.7114	&0.7274&	0.7176 &0.7250	&0.7423	&0.7317\\
\hline
\hline
\textbf{Weighted Voting-based Ensemble}	&0.7077&	0.7017&	0.7012 &0.7222	&0.7175&	0.7164\\
\textbf{Majority Voting-based Ensemble}	&0.7787&	0.7743&	0.7744 &0.7917	&0.7864	&0.7870\\
\hline
\end{tabular}
\label{tab-med-macro}
\end{table*}

\subsubsection{Medication Identification}

Since medication identification can be viewed as binary classification problems including two classes, namely, ``drug" and ``"non-drug", we applied precision, recall and F-score to evaluate the performance. Table~\ref{table-med} presented the performance comparison on medication identification between pretrained BERT models and proposed ensemble models. Similar observations are obtained based on results in Table~\ref{tab-med-micro} and~\ref{tab-med-macro} that pretrained BERT models obtained promising performance based on evaluation results,. In addition, In addition, BERT pretrained on biomedical data can achieve higher performance, for example, regarding $BioBERT$ performance, which indicated that the biomedical data can further contributed more than general data.   However, the performance on medication event classification were lower than that of  medication identification since this subtask needs more complex context to complete classification. 

Moreover, regarding the proposed models, majority voting-based ensemble models built based on BERT achieved optimal performance, which means that integrating multiple pretrained BERT models can further improve the performance. For instance, Precision can be improved by 6\% while Fscore was increased by 2\% for strict evaluation. However, BERT-based weighted ensemble cannot enhance performance due to inappropriate weights assigned to the pretrained BERT models. 

In summary, in terms of observations aforementioned, the proposed ensemble models can enhance the performance for medication event classification through evaluating the results with various evaluation metrics. It means that integrating multiple predictions from different pretrained BERT can leverage complementary information to enhance performance effectively. 

\begin{table*}[!ht]
\caption{Performance comparison on medication identification between pretrained BERT models and proposed ensemble models}
\centering
\begin{tabular}{|l|ccc|ccc|}
\hline
\textbf{Model}&\multicolumn{3}{c|}{\textbf{Strict Evaluation Performance}}&\multicolumn{3}{c|}{\textbf{Lenient Evaluation Performance}} \\
\hline
\textbf{Pretrained BERT}  & \textbf{Precision} & \textbf{Recall} & \textbf{Fscore} & \textbf{Precision} & \textbf{Recall} & \textbf{Fscore} \\
\hline
$BERT_{base}$	&0.8493  & 0.7879 &0.8181 &	 0.8791 &0.8169	&0.8469\\
$BioBERT_{pmc}$	&0.8521	&0.7971	&0.8237 & \textbf{0.8830}		&0.8260	&0.8535\\
$BioBERT_{pubmed}$&	0.8508&	0.8084	&0.8291 & 0.8759	&0.8322&	0.8585\\
$BioBERT_{pubmed\_pmc}$	&\textbf{0.8669}	&0.7965&	0.8309 & 0.8813	&0.8388	&0.8552\\
$Clinical BERT$	&0.8421	&0.8135	&0.8276 &0.8685	&0.8390	&0.8535\\
$Discharge BERT$	&0.8507	&0.8082	&0.8287 &0.8788	&0.8345	&0.8561\\
$BioClinical BERT$	&0.8525&	0.8229	&0.8373 &0.8702&	0.8396	&0.8546\\
$BioDischarge BERT$	&0.8381	&0.8201	&0.8241 &0.8710&	0.8418	&0.8572\\
$BioReddit BERT$	&0.8379	&\textbf{0.9256}	&0.8802 &0.8670&	0.9609	&0.9115\\
$Roberta_{base}$	&0.8382	&0.9233	&0.8783 &0.8650	&0.9518	&0.9063\\
$Roberta_{large}$	&0.8484	&0.9235&	\textbf{0.8844} &0.8724	&0.9495	&\textbf{0.9193}\\
\hline
\hline
\textbf{Weighted Voting-based Ensemble}	&0.8469	&0.9252	&0.8843&0.8708&	0.9512	&0.9092\\
\textbf{Majority Voting-based Ensemble}	&\textbf{0.9365}	&0.8704	&\textbf{0.9092} &\textbf{0.9580}	&0.8904	&\textbf{0.9230}\\
\hline
\end{tabular}
\label{table-med}
\end{table*}

\section{Related Work}
\label{sec3}

Electronic Health Records (EHRs) refer to systematically recorded details and history of a patient. It is a necessary part of the medical sector nowadays because of its reliability and flexibility. Previously most of the health records were recorded and stored in papers and this is a problem considering the cost and the fickle nature of these data as any accident could contribute to a loss of millions patient’s records. To battle these issues, EHRs were introduced into the medical system where all the history of a patient would be updated systematically.  EHRs generated nowadays is a huge source of data but the problem with the EHR data is that it’s very unstructured as they are generated in real time containing a patient’s medical history. 

Due to the unstructured nature of EHRs data, using any machine learning algorithm on EHRs is not efficient. Previous research has shown that due to the lack of frequency of appearance of the medical and biomedical terms in spoken and written language, conventional deep learning models such as Recurrent Neural Networks (RNN), Long Short-term Memory (LSTM) don’t perform well\cite{bilstm}. In addition, regarding the unstructured nature of the EHRs data, pre-processing and fitting the data into a machine learning model becomes challenging. Most of the traditional machine learning algorithms perform poorly in clinical and biomedical field because of the unstructured nature of health records and their nature of complexity. In the field of clinical and biomedical, the sensitivity of data contributes to a shortage of data that is required to apply a deep learning model. On the other side, EHRs contain a patient’s name, age, address, history of disease, family’s details. This sort of information is very private and very prone to exploitation. As a result, biomedical and medical data are scarce. Using deep learning models such as BERT on the EHRs data does not contribute to a good performance because of the sheer lack of the availability of data. Moreover, simply using deep learning models as transfer learning on the EHR dataset also contribute to limited performance because of the lack of the appearance of the biomedical and medical terms in written and spoken language\cite{bert}. That is why pre-training these deep learning models with the publicly available medical and biomedical data can contribute to a better performance on EHR data and this has been proved by published research\cite{biobert}\cite{elmo}\cite{cbert}.



\section{Conclusion}
\label{sec7}

%
%

This paper proposed BERT-based ensemble models for medication event recognition involved in N2C2 2022 competition. It employed multiple pretrained BERT models to perform the medication event recognition on CMED training datasets, where these pretrained BERT models were fine-tuned on CMED testing datasets. The prediction results were evaluated with strict and lenient metrics based on Micro-average and Macro-average precision, recall and Fscore. Evaluation results demonstrated that the proposed model can improve the performance significantly.

Future work will perform more investigation on how to enhance weighted ensemble models regarding their performance since the weighted ensemble models can help figure out which pretrained BERT model can contribute the results more than others in the ensemble process, which helps us interpret ensemble predictions.

\section*{Acknowledgment}
\label{acknowledgement}
\textcolor{red}{This research work is supported in part by the U.S. Office of the Under Secretary of Defense for Research and Engineering (OUSD(R\&E)) under agreement number FA8750-15-2-0119. The U.S. Government is authorized to reproduce and distribute reprints for governmental purposes notwithstanding any copyright notation thereon. The views and conclusions contained herein are those of the authors and should not be interpreted as necessarily representing the official policies or endorsements, either expressed or implied, of the Office of the Under Secretary of Defense for Research and Engineering (OUSD(R\&E)) or the U.S. Government.}

\ifCLASSOPTIONcaptionsoff
  \newpage
\fi


\bibliographystyle{ieeetr}
\bibliography{references}
\end{document}